\documentclass[conference]{IEEEtran}
\IEEEoverridecommandlockouts

\usepackage{graphicx}
\usepackage{algorithm}
\usepackage{stackengine}
\usepackage{cite}
\usepackage{color}
\usepackage{xcolor}
\usepackage{amsmath,amsthm,amssymb,amsfonts,bm}
\usepackage{graphicx}
\usepackage{subcaption} 
\usepackage{textcomp}
\usepackage{multirow}
\usepackage{url}
\usepackage{graphicx}
\usepackage{subfig}
\usepackage{caption}

\def\BibTeX{{\rm B\kern-.05em{\sc i\kern-.025em b}\kern-.08em
    T\kern-.1667em\lower.7ex\hbox{E}\kern-.125emX}}

\usepackage[noend]{algpseudocode}

\makeatletter
\newcommand*{\rom}[1]{\expandafter\@slowromancap\romannumeral #1@}

\def \v x{\bm x}

\def \v x{\bm X}

\renewcommand{\v}[1]{\ensuremath{\boldsymbol{#1}}}

\begin{document}

\title{Byzantine-Resilient Federated Learning via Distributed Optimization
}

\author{\IEEEauthorblockN{Yufei Xia}
\IEEEauthorblockA{ 
\textit{Institut Polytechnique de Paris}\\
France \\
yufei.xia@ip-paris.fr}
\and
\IEEEauthorblockN{Wenrui Yu}
\IEEEauthorblockA{
\textit{Aalborg University}\\
Denmark \\
wenyu@es.aau.dk}
\and
\IEEEauthorblockN{Qiongxiu Li}
\IEEEauthorblockA{\textit{Aalborg University}\\
Denmark \\
qili@es.aau.dk}
}

\maketitle

\begin{abstract}
Byzantine attacks present a critical challenge to Federated Learning (FL), where malicious participants can disrupt the training process, degrade model accuracy, and compromise system reliability. Traditional FL frameworks typically rely on aggregation-based protocols for model updates, leaving them vulnerable to sophisticated adversarial strategies. 
In this paper, we demonstrate that distributed optimization offers a principled and robust alternative to aggregation-centric methods. Specifically, we show that the Primal-Dual Method of Multipliers (PDMM) inherently mitigates Byzantine impacts by leveraging its fault-tolerant consensus mechanism. Through extensive experiments on three datasets (MNIST, FashionMNIST, and Olivetti), under various attack scenarios including bit-flipping and Gaussian noise injection, we validate the superior resilience of distributed optimization protocols. Compared to traditional aggregation-centric approaches, PDMM achieves higher model utility, faster convergence, and improved stability. Our results highlight the effectiveness of distributed optimization in defending against Byzantine threats, paving the way for more secure and resilient federated learning systems.
\end{abstract}


\section{Introduction}
Federated learning (FL) enables collaborative model training across distributed clients without exposing raw data \cite{mcmahan2017communication,li2020federated}. By design, FL addresses privacy concerns inherent in centralized data collection, making it particularly appealing for critical applications such as healthcare, IoT, and edge computing.   A typical FL workflow involves three steps: 1) initializing a global model, 2) local client updates, and 3) aggregating updates into a shared model.  Existing FL protocols can be broadly categorized along two axes: topology and protocol \cite{li2025centralized}. By topology, FL can be centralized (CFL) or decentralized (DFL). In CFL, a central server aggregates updates using protocols like FedAvg and FedSGD \cite{mcmahan2017communication}, whereas DFL relies on peer-to-peer communication \cite{niwa2020edge,li2024topology}. Despite its advantages in privacy preservation, FL remains vulnerable to privacy attacks~\cite{geiping2020inverting,yin2021see,yang2022using,li2024perfect}, where adversaries aim to reconstruct sensitive data from shared model updates.  While the privacy implications of various FL protocols have been extensively studied~\cite{li2024eusipco,pasquini2023security,yu2024privacy,ji2024re}, their robustness against Byzantine attacks remains underexplored. 

Byzantine attacks, where malicious clients submit falsified updates to degrade or manipulate the global model, pose a critical threat to FL systems\cite{so2020byzantine, praneeth2020byzantine}. Current literature predominantly focuses on attacks targeting aggregation-based protocols like FedAvg, where adversaries exploit the server’s reliance on averaging to inject skewed gradients.  Common attack strategies include bit-flipping~\cite{rakin2019bit}, Gaussian noise injection~\cite{wei2020federated,geyer2017differentially}, and label flipping~\cite{moharram2022defending}. To defend against these attacks, robust aggregation rules such as Trimmed Mean~\cite{yin2018byzantine} and Krum~\cite{blanchard2017machine} have been proposed. However, these approaches remain reactive, often relying on outlier detection and assumptions about the redundancy or majority of honest clients. Moreover, they do not fundamentally address the inherent vulnerability of aggregation-centric FL protocols to Byzantine behavior.    An open question remains: are alternative FL protocols, particularly those based on joint optimization, also susceptible to Byzantine attacks? Distributed optimization methods such as the Alternating Direction Method of Multipliers (ADMM)~\cite{boyd2011distributed,giselsson2016linear} and the Primal-Dual Method of Multipliers (PDMM)~\cite{zhang2017distributed,sherson2018derivation,heusdens2024distributed} have demonstrated convergence guarantees and fault tolerance in distributed systems. However, their resilience to Byzantine behavior in FL settings has not been thoroughly investigated.

In this paper, we bridge this gap by systematically analyzing the Byzantine resilience of distributed optimization based FL protocols. Unlike aggregation-based approaches that separate local training from global aggregation, joint optimization protocols synchronize client updates through iterative consensus mechanisms and dual variable constraints. We hypothesize that these mechanisms inherently limit adversarial impact by enforcing global agreement and leveraging redundancy in distributed subproblems.   To validate this,we conduct the first empirical comparison between FedAvg (representing aggregation-based protocols) and PDMM (representing joint optimization) under Byzantine attack scenarios, including label flipping and Gaussian noise injection. Experiments on MNIST, FashionMNIST, and Olivetti datasets demonstrate that PDMM achieves superior accuracy and stability across both CFL and DFL topologies. These results confirm that joint optimization protocols inherently constrain Byzantine impacts.

Our work challenges the prevailing focus on post-hoc defenses for aggregation-based FL and highlights the underexplored potential of protocol-level robustness. By demonstrating that distributed optimization algorithms, such as ADMM and PDMM, naturally mitigate Byzantine failures, we advocate  for rethinking FL protocol design in adversarial environments.

\section{Preliminary}
This section introduces the necessary fundamentals  and the notations used throughout the paper are summarized in Table~\ref{tab:notation}.
\subsection{Aggregation based Federated Learning}
A widely adopted aggreagation protocol in FL, known as \emph{FedAvg}~\cite{mcmahan2017communication}, operates as follows:
\begin{enumerate}
    \item \textbf{Initialization}: At iteration $t=0$, the server randomly initializes the global model weights $\mathbf{w}_s^{(0)}$ and distributes them to all participating nodes.
    
    \item \textbf{Local Update}: At each iteration $t$, each node $i$ receives the current global model $\mathbf{w}_s^{(t-1)}$ and updates its local model $\mathbf{w}_i^{(t)}$ based on its private dataset $\mathbf{x}_i$. The resulting local model updates are then transmitted back to the server.
    
    \item \textbf{Model Aggregation}: Upon receiving the local updates, the server aggregates them to form an updated global model:
    \begin{align}\label{eq.w_ave}
        \mathbf{w}_s^{(t)} = \sum_{i=1}^{N} a_i \mathbf{w}_i^{(t)},
    \end{align}
\end{enumerate}
where $a_i$ is the aggregation weight assigned to client $i$ and $\sum_{i=1}^{N} a_i=1$.  Steps 2 and 3 are repeated until the global model converges or a predefined stopping criterion is satisfied. Another commonly used protocol is \emph{FedSGD}~\cite{mcmahan2017communication}, which assumes full-batch training on the client side and performs similar aggregation strategy.

\subsection{Byzantine Attacks}
Byzantine robustness has been a major research concern in FL due to the presence of malicious or unreliable clients. Byzantine clients can deliberately inject false updates or non-honest data, compromise global model accuracy and delay convergence. In this paper, we focus on two representative and widely studied Byzantine attack models:
\begin{itemize}
    \item \textbf{Bit Flipping Attack}~\cite{rakin2019bit}: Malicious clients invert the sign of their model updates to counteract the contributions of honest clients. Formally, for a Byzantine client $i \in \mathcal{B}$, the update is:
    \begin{equation}
        \hat{\mathbf{w}}_i^{(t+1)} = -\mathbf{w}_i^{(t+1)}
    \end{equation}
    Bit flipping attacks can severely hinder or entirely prevent convergence, particularly when a significant fraction of clients are adversarial.
    \item \textbf{Gaussian Noise Attack}~\cite{wei2020federated,geyer2017differentially}: Malicious clients inject random Gaussian noise into their model updates to introduce instability and disrupt learning:
    \begin{equation}
        \hat{\mathbf{w}}_i^{(t+1)} = \mathbf{w}_i^{(t+1)} + \mathcal{N}(0, \sigma^2)
\end{equation}
Unlike bit flipping, which systematically misguides the optimization process, Gaussian noise injection adds excessive randomness, preventing meaningful learning progress and increasing variance in model updates.
\end{itemize}

These types of Byzantine attacks are particularly disruptive to aggregation-based FL protocols, such as FedAvg and FedSGD, which generally lack mechanisms to defend against malicious updates. Common defense strategies include robust aggregation methods such as Trimmed Mean~\cite{yin2018byzantine} and Krum~\cite{blanchard2017machine}, which aim to filter out outliers and mitigate the influence of compromised clients.

In this work, we demonstrate that distributed optimization methods (e.g., ADMM or PDMM) can offer inherent robustness against Byzantine attacks. By jointly optimizing local updates and enforcing global consensus constraints, A/PDMM limits the ability of malicious clients to degrade learning performance. The following sections provide a detailed description of our methodology and experimental validation. 

\section{Methodology}
This section presents the proposed Byzantine-resilient FL framework based on distributed optimization.   Besides, the attack models are also detailed with formulas. 

\subsection{Distributed Optimization in FL}
Distributed optimization addresses global learning objectives across decentralized networks. The general formulation is expressed as the following constrained optimization problem:
\begin{align}
\label{eq:distributed_optimization}
\begin{array}{ll}
    \min \limits_{\{\mathbf{w}_i \}} & \sum\limits_{i \in \mathcal{N}} f_i(\mathbf{w}_i), \\
    \text{subject to} & \mathbf{B}_{i|j} \mathbf{w}_i + \mathbf{B}_{j|i} \mathbf{w}_j = \mathbf{0}, \quad \forall (i,j) \in \mathcal{E},
\end{array}
\end{align}
The matrices $\mathbf{B}_{i|j}$ define linear edge constraints that ensure consensus among neighboring nodes. Typically, $\mathbf{B}_{i|j} = -\mathbf{B}_{j|i} = \pm \mathbf{I}$ to ensure that all nodes converge to the same model.
\begin{table}[t]
\centering
\caption{Notation Table}
\renewcommand{\arraystretch}{1.2} 
\begin{tabular}{c|l}
\hline
\textbf{Symbol} & \textbf{Description} \\ \hline
$ \mathbf{w}_i $  & Local model parameter at client $i$ \\
$ \mathbf{w}_s $  & Global model parameter at sever \\
$ \mathbf{y}_{i\mid j} $  & Transmission variable in PDMM from client $i$ to $j$ \\
$ \mathcal{B} $  & Set of Byzantine clients \\
$ \mathcal{N} $  & Set of clients \\
$\mathcal{E}$ & Set of edges \\
$ f_i( \mathbf{w}) $  & Loss function at client $i$ \\
$    t   $  & Iteration\\ 
$ c $  & PDMM penalty parameter  \\
$ \eta $  & learning rate  \\
$ \sigma $ & Standard deviation of Gaussian noise\\
$ \hat{(\cdot)} $ & Malicious behavior from Byzantine clients\\
\hline
\end{tabular}
\label{tab:notation}
\end{table}

\subsection{PDMM based FL}
The PDMM algorithm introduces dual variables to enforce consensus and stabilize model updates. Each client optimizes its local model while exchanging information with a central server (in CFL) or directly with peers (in DFL) to maintain consistency. Algorithm~\ref{alg:PDMM} outlines PDMM for CFL. The corresponding version for DFL follows the same structure but restricts communication to neighboring nodes \cite{zhang2022revisiting}.

\begin{algorithm}
\small
\caption{PDMM based CFL}
\label{alg:PDMM}
\begin{algorithmic}[1]
\Require Penalty parameter $c > 0$, number of clients $N$
\State \textbf{Initialize:} Local models $\mathbf{w}_i^{(0)}= \mathbf{w}_{\text{s}}^{(0)}$, dual variables $\mathbf{y}_{i|j}^{(0)}$ for all $i \in \{1, 2, \dots, N\}$ 
\While{No convergence}
    \For{each client $i$ \textbf{in parallel}} \Comment{Primal update}
        \State Compute local model update:
        \[
        \mathbf{w}_i^{(t+1)} = \arg\min_{\mathbf{w}_i} \left[ f_i(\mathbf{w}_i) + \frac{c}{2} \| \mathbf{w}_i - \mathbf{y}_{s|i}^{(t)} \|^2 \right]
        \]
        \State Update auxiliary variable $\mathbf{y}_{i|s}^{(t+1)}$ of clients:
        \[
        \mathbf{y}_{i|s}^{(t+1)} = 2\mathbf{w}_i^{(t+1)} - \mathbf{y}_{s|i}^{(t)}
        \]
        \State \textbf{Server:} Aggregate models \Comment{Second update}
        \[
        \mathbf{w}_{\text{s}}^{(t+1)} = \frac{1}{N} \sum_{i=1}^{N} \mathbf{y}_{i|s}^{(t+1)}
        \]
        \State Update auxiliary variable $\mathbf{y}_{s|i}^{(t+1)}$ of server:
        \[
            \mathbf{y}_{s|i}^{(t+1)} = 2\mathbf{w}_s^{(t+1)} - \mathbf{y}_{i|s}^{(t)}
        \]
        \State $t = t + 1$
    \EndFor
\EndWhile
\end{algorithmic}
\end{algorithm}
 
 The server is denoted by $s$ and $\mathbf{w}_s$ is the model of server. The server will initialize the model for each node $i$ as the same in FedAvg. The joint optimization aggregation of PDMM is divided into two steps. Each client $i$ performs a local model update by minimizing the loss function, which includes a term involving the dual variables of server $\mathbf{y}_{s|i}$ to regulate the local model update, where c is the regulation parameter. The update is given by:
\begin{equation}
    \mathbf{w}_i^{(t+1)} = \arg\min_{\mathbf{w}_i} \left[ f_i(\mathbf{w}_i) + \frac{c}{2} \| \mathbf{w}_i - \mathbf{y}_{s|i}^{(t)} \|^2 \right]\\
    \label{local update of PDMM}
\end{equation}

After the local model update, each client updates its dual variable $\mathbf{y}_{i|s}$ based on its local model:
\begin{equation}
    \mathbf{y}_{i|s}^{(t+1)} = 2\mathbf{w}_i^{(t+1)} - \mathbf{y}_{s|i}^{(t)}
    \label{local update of PDMM}
\end{equation}

Then the server aggregates the models from all clients by computing the average:
\begin{equation}
    \mathbf{w}_{\text{s}}^{(t+1)} = \frac{1}{N} \sum_{i=1}^{N} \mathbf{y}_{i|s}^{(t+1)}
    \label{local update of PDMM}
\end{equation}

At the end of each round, the server updates the global dual variables based on the aggregated models:
\begin{equation}
    \mathbf{y}_{s|i}^{(t+1)} = 2\mathbf{w}_s^{(t+1)} - \mathbf{y}_{i|s}^{(t)}
    \label{local update of PDMM}
\end{equation}

\subsection{Analysis of Byzantine robustness}
PDMM offers inherent robustness against Byzantine attacks through its joint optimization framework and consensus enforcement mechanism. By integrating local updates with global consistency constraints, PDMM mitigates the influence of malicious clients without relying on external defense mechanisms. Unlike aggregation-based methods such as FedAvg and FedSGD, which are vulnerable to adversarial manipulation during the model aggregation step, PDMM stabilizes local updates by penalizing deviations from consensus through dual variable updates. This structure inherently limits the ability of Byzantine clients to disrupt the learning process.

Theoretical analyses in~\cite{jonkman2018quantisation,liang2016convergence} demonstrate that PDMM can tolerate finite perturbations while maintaining convergence guarantees. Hence, as long as the number of attack rounds or the magnitude of injected noise remains bounded, the algorithm ensures the integrity and stability of the global model.
Hence, PDMM achieves resilience through its core optimization process, making it a principled and effective solution for Byzantine-resilient FL.

\begin{figure}[h]
    \centering

    \begin{subfigure}{0.50\textwidth}
        \centering
        \includegraphics[width=\textwidth]{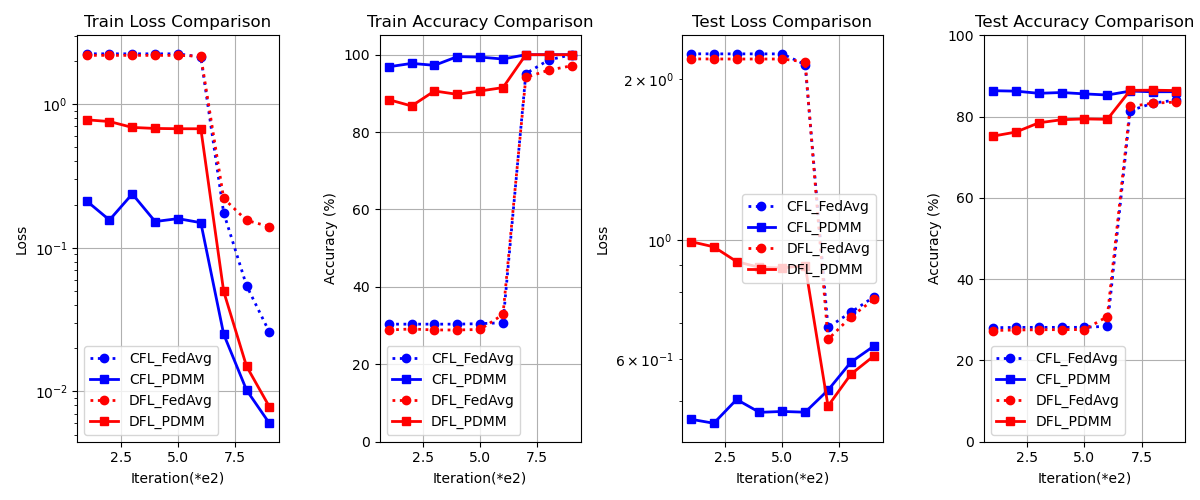}
        \caption{ MNIST}
    \end{subfigure}
    \vspace{0.1cm} 
    \begin{subfigure}{0.50\textwidth}
        \centering
        \includegraphics[width=\textwidth]{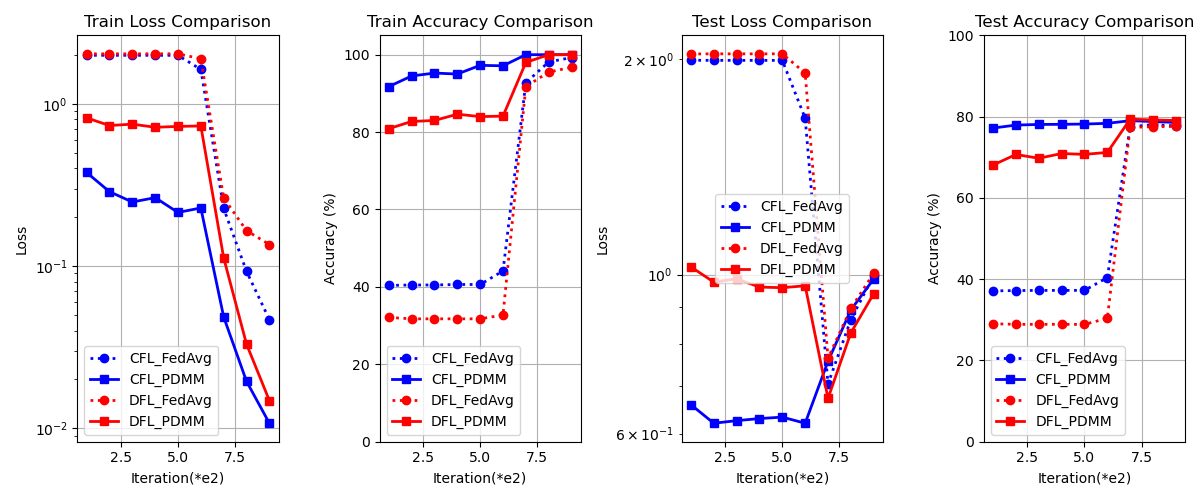}
         \caption{FashionMNIST}
    \end{subfigure}
        \vspace{0.1cm} 
    \begin{subfigure}{0.50\textwidth}
        \centering
        \includegraphics[width=\textwidth]{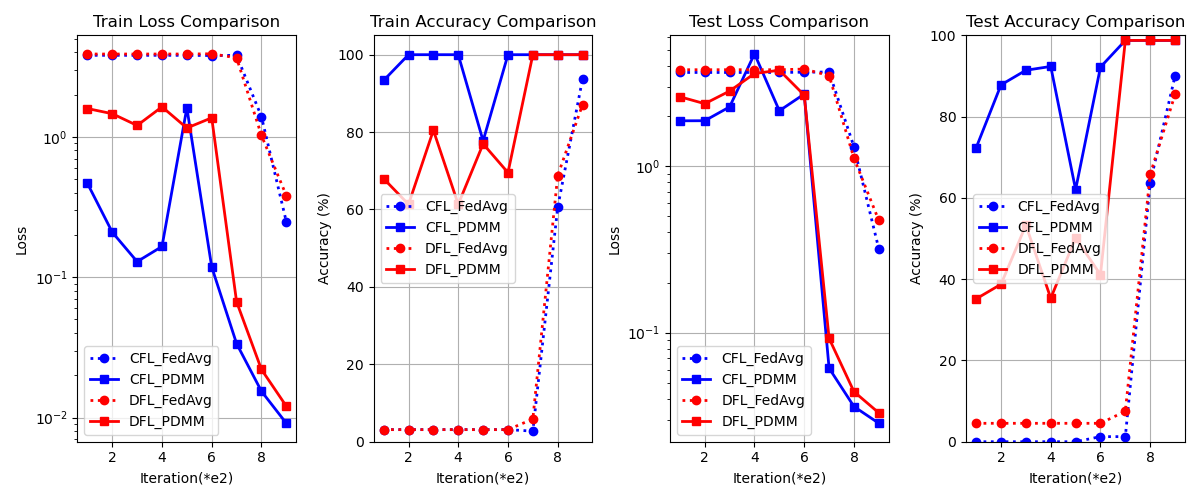}
       \caption{Olivetti}
    \end{subfigure}
            \caption{Byzantine robustness comparison of PDMM based FL protocols over traditional FedAvg based protocols against big flipping attack over three datasets and two types of topologies (CFL v.s. DFL).}
                \label{fig:bitflipping} 
\end{figure}

\begin{figure}[h]
    \centering
    \begin{subfigure}[b]{0.50\textwidth}
        \centering
        \includegraphics[width=\textwidth]{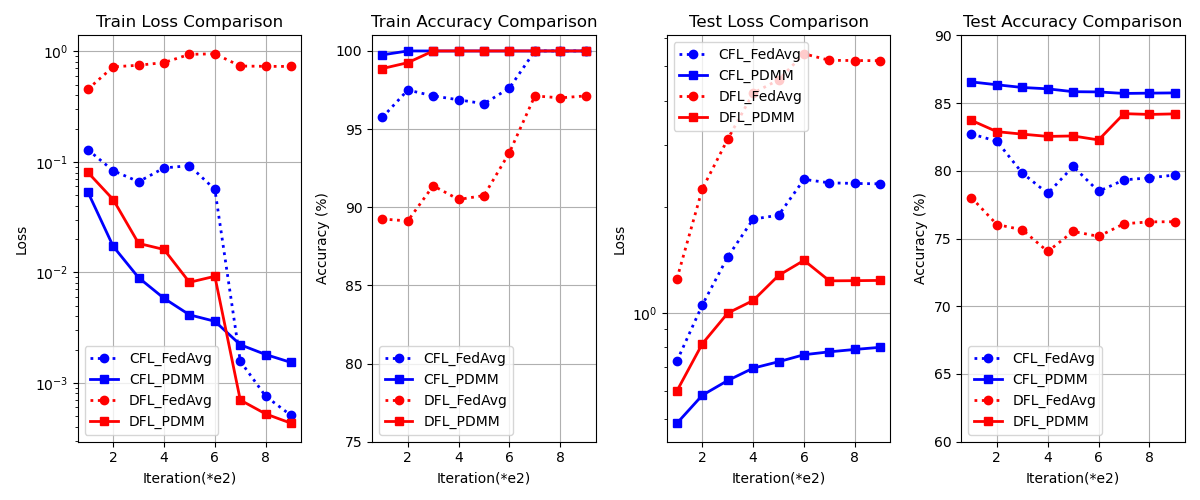}
         \caption{ MNIST}
    \end{subfigure}
    \begin{subfigure}[b]{0.50\textwidth}
        \centering
        \includegraphics[width=\textwidth]{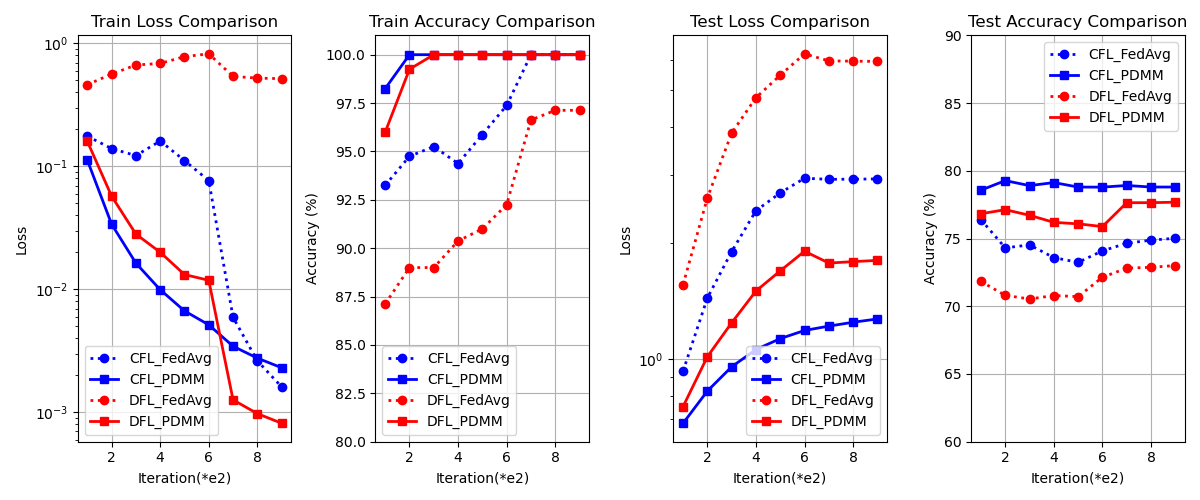}
        \caption{ FashionMNIST}
    \end{subfigure}
    \begin{subfigure}[b]{0.50\textwidth}
        \centering
        \includegraphics[width=\textwidth]{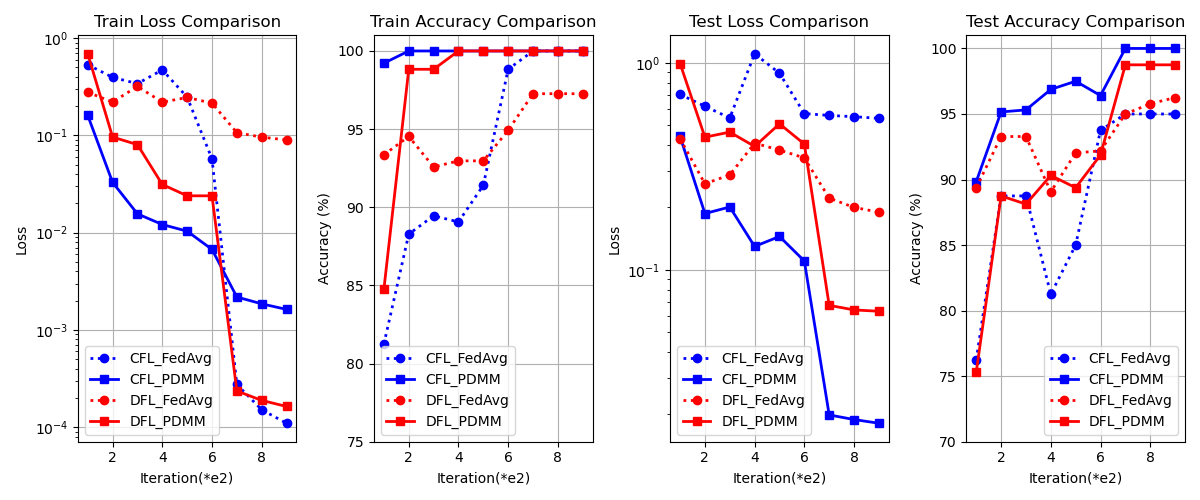}
        \caption{Olivetti}
    \end{subfigure}
            \caption{Byzantine robustness comparison of PDMM based FL protocols over traditional FedAvg based protocols against Gaussian noise attack over three datasets and two types topologies (CFL v.s., DFL).}
            \label{fig:gaussiannoise}       
\end{figure}

\section{Numerical Results}

In this section, we present empirical evaluations that demonstrate the effectiveness of distributed optimization-based federated learning protocols in mitigating Byzantine attacks. Specifically, we compare distributed optimization methods—such as the Primal-Dual Method of Multipliers (PDMM)—against traditional aggregation-based federated learning frameworks under various adversarial conditions.  

\subsection{Experimental setup}
We evaluate the robustness of PDMM and FedAvg against 2 Byzantine attacks by conducting experiments on MNIST \cite{lecun1998gradient}, FashionMNIST \cite{xiao2017fashion}, and Olivetti \cite{olivetti_faces} datasets. Olivetti includes a set of 400 facial images of 40 people. The used network is a 2 layer multilayer perception (MLP) structure. Experiments involve 10 clients, with 2 Byzantine nodes launching attacks in the first 600 rounds over 1000 training rounds. The data is independently and identically distributed across each client.  We consider a random geometric graph of $n=10$ nodes~\cite{dall2002random} as the DFL topology. In each iteration, we use 10 steps of gradient descent for both CFL and DFL. For MNIST and FashionMNIST, the learning rate is set as $\eta=0.05$ and $\sigma=0.1$; for Olivetti, we set $\eta=0.04$ and $\sigma=0.2$.

We consider four FL architectures: (1) CFL with PDMM (centralized, joint optimization), (2) CFL with FedAvg (centralized, separate aggregation), (3) DFL with PDMM (decentralized, joint optimization), and (4) DFL with FedAvg (decentralized, separate aggregation). The decentralized version of FedAvg follows the same procedure as in the centralized setting but replaces the global server aggregation with local neighborhood aggregation, which we also refer to as FegAvg in our comparisons.

\subsection{Performance comparison under bit flipping Attacks} 
Figure~\ref{fig:bitflipping} presents the performance comparison between PDMM-based protocols (solid lines) and FedAvg-based protocols (dotted lines)  under bit-flipping attacks. In this scenario, two malicious nodes execute bit-flipping attacks across 600 out of 1000 attack rounds. 

From the plots, it is evident that distributed optimization protocols (denoted as DFL\_PDMM and CFL\_PDMM) consistently outperform aggregation-based methods (CFL\_FedAvg and CFL\_FedAvg). Specifically, DFL\_PDMM achieves substantially lower training and test loss, while maintaining higher training and test accuracy across all three datasets. For instance, under bit-flipping attacks in CFL configuration, PDMM outperforms FedAvg by 37\% on FashionMNIST and 56\% on MNIST for test accuracy. As for test accuracy of DFL structure, PDMM is 40\% higher for FashionMNIST and 37\% higher for MNIST than FedAvg. The results demonstrate that distributed optimization protocols inherently mitigate the detrimental effects of Byzantine nodes, ensuring stable convergence and robust performance even under sustained adversarial interference.

\subsection{Performance comparison under Gaussian noise attacks}
Figure~\ref{fig:gaussiannoise} shows the results for the same FL systems under Gaussian noise injection attacks. Again, two malicious nodes perform attacks over 600 out of 1000 rounds. As illustrated in the figure, distributed optimization protocols outperform aggregation-based methods in both convergence speed and final model utility. DFL\_PDMM exhibits remarkable stability, with significantly reduced variance in both accuracy and loss metrics. Under Gaussian noise attack, the test accuracy of PDMM is 8.75\% higher than FedAvg for CFL structure with Olivetti datasets. Even under intense Gaussian noise perturbations, the distributed optimization framework consistently achieves superior robustness and convergence compared to traditional aggregation-centric federated learning protocols.

\subsection{Performance superiority is independent of topology type}
A key observation from our experiments is that the superiority of distributed optimization-based federated learning protocols is largely independent of the network topology. Whether implemented in CFL or DFL settings, distributed optimization protocols consistently outperform aggregation-based approaches.

This robustness is attributable to the inherent fault-tolerant design of distributed optimization methods like PDMM. By distributing computation and information exchange more equitably among nodes, and by mitigating the influence of any single malicious actor, these protocols reduce the risk posed by Byzantine behaviors. Our empirical results clearly demonstrate that across varying topologies and attack strategies, distributed optimization remains the more resilient and effective approach.

\section{Conclusion}
In this paper, we demonstrate that distributed optimization protocols, such as PDMM, offer inherent robustness against Byzantine attacks in FL. Unlike traditional aggregation-based methods, PDMM enforces consensus through joint optimization, effectively limiting the impact of malicious clients. Our empirical results, validated across diverse datasets and attack scenarios, show that PDMM consistently outperforms traditional aggregation-based protocols like FedAvg in both centralized and decentralized FL architectures. Notably, PDMM achieves higher accuracy, faster convergence, and greater stability under common Byzantine attacks, including bit-flipping and Gaussian noise injection. These findings highlight the potential of protocol-level robustness and suggest a shift away from reactive defenses in FL. 
\bibliographystyle{IEEEbib}
\bibliography{refs}

\end{document}